# Heuristic Dynamic Programming for Adaptive Virtual Synchronous Generators

Sepehr Saadatmand, Mohammad Saleh Sanjarinia, Pourya Shamsi, Mehdi Ferdowsi, and Donald C. Wunsch
Department of Electrical and Computer Engineering
Missouri University of Science and Technology
sszgz@mst.edu, mswvq@mst.edu, shamsip@mst.edu, ferdowsi@mst.edu, dwunsch@mst.edu

*Abstract*—In this paper a neural network heuristic dynamic programing (HDP) is used for optimal control of the virtual inertia-based control of grid connected three-phase inverters. It is shown that the conventional virtual inertia controllers are not suited for non-inductive grids. A neural network–based controller is proposed to adapt to any impedance angle. Applying an adaptive dynamic programming controller instead of a supervised controlled method enables the system to adjust itself to different conditions. The proposed HDP consists of two subnetworks: critic network and action network. These networks can be trained during the same training cycle to decrease the training time. The simulation results confirm that the proposed neural network HDP controller performs better than the traditional direct-fed voltage an/or reactive power controllers in virtual inertia control schemes.

*Index Terms*-- grid connected inverter, heuristic dynamic programming, neural network, virtual synchronous generator

## I. INTRODUCTION

Thanks to the growing demand for more sustainable generation, distributed energy resources (RESs) are replacing the traditional generations. Consequently, utilization of inertia-less power electronic inverters has led to a dramatic fall in system inertia especially in islanded microgrids. The reason is that RESs, such as photovoltaics (PVs) and wind turbines, are tied to the grid using fast-response power electronic converters, which do not have any inertia. The stability of the system and the voltage and frequency fluctuation increase by the decrease in system inertia, which is the result of the increasing RES penetration [1].

To cope with this concern, several solutions have been proposed. Considering the fact that power system inertia is essentially provided by the large kinetic energy buffered in synchronous generators, the concept of virtual synchronous generator/machine (VSG/VSM) was introduced recently to virtually imitate the response of the traditional synchronous generators virtually to interact with the power system to improve system inertia, resiliency, microgrid stability, and output impedance [2]-[4]. To improve frequency stability and solve the control problem of microgrid frequency, several control methods have been applied to control VSGs. A traditional proportional controller (PI) has been used for virtual inertia control and its application in wind power [5]. However, recent studies show that these controllers have inferior performance in particular when facing uncertainties concerning the angle of grid's impedance [some references that have used other controllers].

Neural networks are used in many areas such as image processing, speech recognition, text mining [6] and control field of study. Dynamic programing (DP) and approximate/adaptive dynamic programing (ADP) can be used for optimal control problem. Combining parametric structure with incremental optimization forms a new class of ADP called adaptive critic design (ACD) that approximate the optimal cost [7]-[8]. Heuristic dynamic programing (HDP) is the simplest form of ACDs and has been studied in different applications, such as vector control of a grid-connected converter [9], power system stability [10], and permanent magnet synchronous motor drive [11].

The main contribution of this paper is to present a new method for developing a neural network-based HDP combined with the virtual inertia strategy for controlling a grid-tied inverter. First, a brief review of virtual inertia control concept, frequency control, and the limitation associated with conventional control for VSGs is presented in Section II. The heuristic dynamic programming concept, its subnetworks, and the implementation process in VSGs are explained in Section III. Lastly, the performance of the proposed HDP controller is evaluated in Section IV.

## II. PRINCIPLE AND MODEL OF A VSG

In this section, the VSGs controller structure and the power flow equation are explained. Moreover, it is shown that the virtual inertia controller performance extremely depends on the grid parameters.

### A. VSG controller

The control block diagram of the VSG system is illustrated in Figure. 1. In this paper, it is assumed that the inverter is connected to a regular dc source. Hence, the inverter can route power from the dc source to inject additional power to the grid during transients (not applicable to solar photovoltaic inverters). The control scheme in Figure. 1 is based on the virtual inertia method. In this model, $X_F$ , $X_L$ , and $R_L$ are the inverter output filter reactance, the inverter to the grid line reactance, and the inverter to grid line resistance respectively.

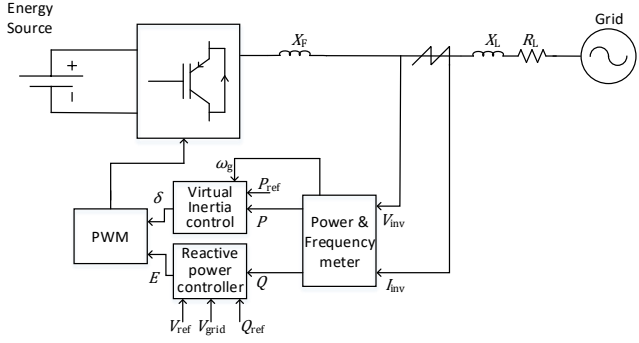

Figure 1. Conventional Virtual Synchronous Generator (VSG) block diagram

The most important component when modelling VSGs is the swing equation of a synchronous generator as follows:

$$P_{in} - P_{out} = J\omega_i \left(\frac{d\omega_i}{dt}\right) + D\Delta\omega \quad (1)$$

where $P_{in}$, $P_{out}$, $\omega_i$, $J$, and $D$ are the input power to VSG, the electric output power, the angular velocity of the virtual rotor, the virtual rotor moment of inertia, and the droop coefficient, respectively. In this equation, $\Delta\omega$ is defined by $\Delta\omega = \omega_i - \omega_g$ where $\omega_g$ is the grid angular velocity while the inverter is connected to the grid or the reference angular velocity while the inverter works in a standalone mode.

The command signal to the inverter includes two parts. First, it needs the RMS value of the inverter voltage or peak value of inverter phase voltage ( $E$ ). Secondly, it needs the inverter power angle with respect to the grid ($\delta$). In order to compute E, the electrical output power can be computed by measuring the inverter voltage signals and the current signal injected into the grid. Having all the necessary parameters in Equation (1), $\Delta\omega$ can be computed at each control cycle. Then, the mechanical phase can be calculated by integrating this frequency.

$$\delta = \int \Delta\omega \cdot dt$$

In high voltage power systems, in order to tune the inverter voltage the reactive power controller with a voltage droop is applied. Applying a voltage droop and an integrator controller provides the RMS/peak value of the voltage using the following equation:

$$E = \frac{1}{K_i} \int \Delta Q \cdot dt - D_v \Delta V$$

where $K_i$ and $D_v$ are the integrator coefficient and the voltage droop, respectively. The inverter reactive power tracking error is given by $\Delta Q = Q_{ref} - Q_e$, and the inverter voltage tracking error is given by $\Delta V = V_{ref} - V_i$. The reference reactive power for the inverter is set to $Q_{ref}$ and the inverter output reactive power can be computed by a power meter block. The variable $V_i$ is the inverter output voltage, and $V_{ref}$ is the reference voltage for the inverter. Figure. 2 shows the active and reactive power controller block diagrams.

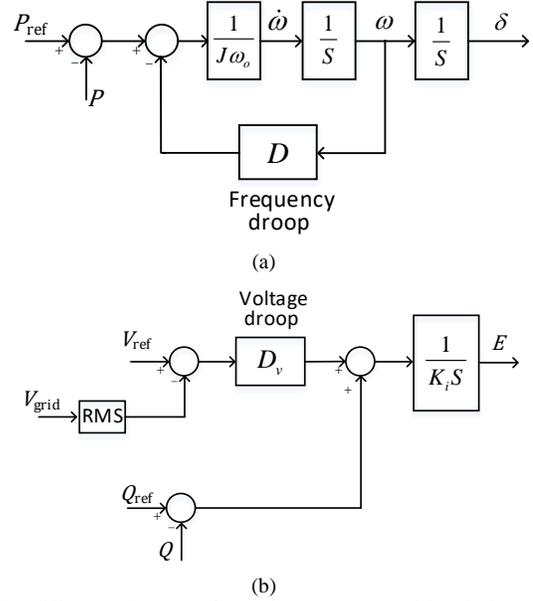

Figure 2. VSG controllers. (a) Power-frequency control block diagram. (b) Voltage-reactive power control block diagram

B. *Power flow equation for grid-connected inverters*

The proposed VSG average model can be derived based on a voltage source as shown in Figure. 3. In the figure, $X_{eq}$ is the equivalent reactance per phase (line and filter) and can be computed as $X_{eq} = X_F + X_L$ , $R_{eq}$ presents the equivalent resistance per phase (line and filter) given as $R_{eq} = R_L$, and $Z_{eq}$ is the equivalent impedance per phase (line and filter) given as $Z_{eq} = jX_{eq} + R_{eq}$. The active and reactive power delivered by the converter to the grid can be expressed as

$$P = \frac{1}{2}\left[\left(\frac{E^2}{Z_{eq}^2} - \frac{EV\cos\delta}{Z_{eq}^2}\right)R_{eq} + \frac{EV}{Z_{eq}^2}X_{eq}\sin\delta\right]$$

$$Q = \frac{1}{2}\left[\left(\frac{E^2}{Z_{eq}^2} - \frac{EV\cos\delta}{Z_{eq}^2}\right)X_{eq} - \frac{EV}{Z_{eq}^2}R_{eq}\sin\delta\right]$$

where $P$ and $Q$ are the delivered active and reactive power (per phas), V is the peak value of the phase voltage of the grid, E is the peak value of the output voltage of the inverter and $\delta$ is the phase angle between the grid voltage and the inverter voltage. For an inductive equivalent impedance (i.e. $X_{eq} \gg R_{eq}$ as in [12], [13],) the active and reactive power can be estimated as:

$$P \approx \frac{EV}{2X_{eq}} \sin\delta \quad (2)$$

$$Q \approx \frac{E}{2X_{eq}}(E - V\cos\delta) \quad (3)$$

Generally, the inverter power angle $\delta$ is small, and $\sin\delta$ can be approximated by $\delta$, and $\cos\delta$ can be approximated by one. Therefore, (2) and (3) can be written as

$$P \approx \frac{EV}{2X_{eq}}\delta \quad (4)$$

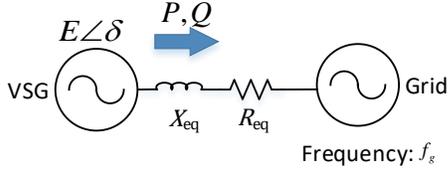

Figure. 3. Equivalent circuit diagram of a grid connected VSG

$$Q \approx \frac{E}{2X_{eq}}(E - V) \quad (5)$$

Equation (4) and (5) clarify that in inductive grids, the active power is proportional to the power angle and the reactive power is proportional to the inverter voltage. In this case, the conventional VSG controller performance is acceptable; nonetheless, in low-voltage grids that are mostly resistive or semi-resistive, this assumption is no longer valid. In other words, Q depends on both the power angle and the voltage magnitude. In order to turn the reactive power controller for non-inductive grids, all the parameters of the system model need to be known to make it possible to design an acceptable reactive power controller. However, in the power system, the inverter might face uncertainties such as line impedance changes or nonlinear behaviors (e.g. transformer saturation) in electrical element that alter the reactive power equation. In this paper, an adaptive dynamic controller, capable of adjusting its parameter, is used to find the optimal solution and the results are compared with the conventional controller performance.

## III. HEURISTIC DYNAMIC PROGRAMING

The integral/proportional integral controller for virtual inertia is limited in terms of the following characteristics. First, the integrator (and the proportional) parameters need to be tuned online based on the system response or the designer experience. In addition, the conventional controllers such as integral controller or proportional controller are designed for linear systems, and it cannot perform well in non-linear systems. In order to tune the coefficients, the linearized model is used. Hence, when the operating point changes, the system model will change and the PI parameters are no longer optimal. Finally, the proposed proportional integral controller is a single-input single-output (SISO) system; however, for the case of a non-inductive grid, both active and reactive error signals are needed as the voltage controller input. The proposed controller in this paper is based on a direct action heuristic dynamic programing scheme to address the aforementioned issues.

Adaptive critic designs are neuro control (neural network–based controller) designs that are suitable for optimal control over time under conditions of uncertainty and noise. [14] proposed ACDs as a new optimization method inspired by approximate dynamic programing and reinforcement learning.

Most important ACDs can be listed as follows: heuristic dynamic programing (HDP), dual heuristic programing (DHP), global heuristic dynamic programing (GHDP), and global dual heuristic programing (GDHP). Typical ACDs consist of two subnetworks: the action neural network and the critic neural network. The main difference between the types of ACDs can be defined in the networks output signal and the error feedback signal.

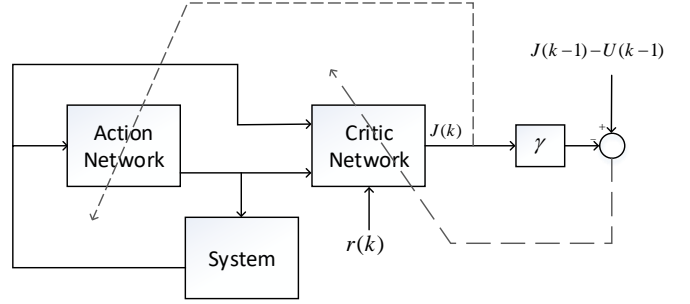

Figure. 4. Heuristic dynamic programing block diagram

HDP can be introduced as the most straightforward form of ADCs. The design scheme of HDP is illustrated in Figure. 4. In HDP, the critic networks learn to approximate the strategic utility or the cost-to-go function, which in dynamic programing is defined as the Bellman's equation function. The action network generates the control signal and feeds it to the system and to the critic network. The state vector and the control signal feed the critic network when its goal is to estimate the cost-to-go function. In Figure. 4, the dashed lines provide the corresponding error signal for the learning procedure of the action and the critic neural network.

### A. Critic neural network

The cost-to-go J function of dynamic programing in Bellman's equation is as follows

$$J(k) = \sum_{k=0}^{\infty} \gamma^i U(k+1)$$

where $\gamma$ is a discount factor, $(0 < \gamma < 1)$, to make sure that the cost to go is bounded, and $U$ is the utility function. The utility function in this paper is defined as

$$U(k) = \sqrt{K_P e_P^2 + K_Q e_Q^2 + K_f e_f^2}$$

where $e_P, e_Q, e_f$ are the error signal fo active power, reactive power, and frequency respectively defined as:

$$e_P = P_{set} - P$$
$$e_Q = Q_{set} - Q$$
$$e_f = f_g - f$$

,

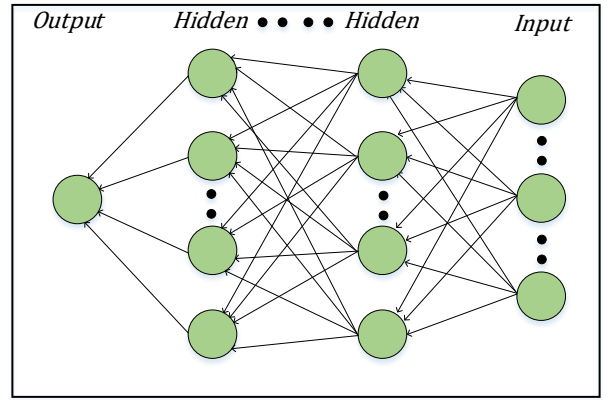

Figure. 5. Fully connected feedforward neural network

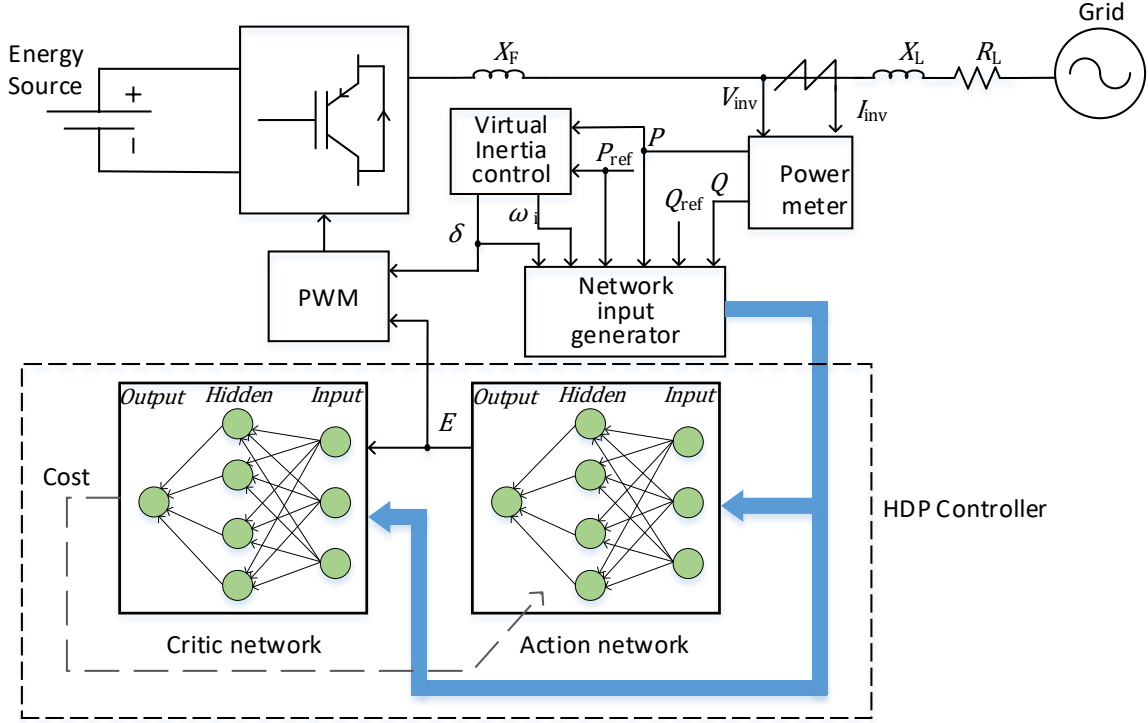

Figure. 6. Block diagram of the proposed controller for VSG

and $K_P$, $K_Q$, $K_f$ are the active power coefficient, the reactive power coefficient, and the frequency coefficient, respectively. These coefficients can also be defined as the weight matrix in a normalized function to define the importance of each error signal. Figure. 5 illustrates the critic network, which is a fully connected multi-layer forward network, consisting of two hidden layers with eight nodes. The input to the neural network is a vector consisting of the inverter active and reactive power, the active power error, the reactive power error, the frequency error, the inverter phase angle, and the action network output. The critic neural network input vector can be expressed as

$$IN_{\{critic\ network\}} = [P\ Q\ e_p\ e_q\ e_f\ \theta_i\ E].$$

Due to the real-time operation of the system, the critic neural network is trained forward in time to minimize the error signal measured over time, which is defined in a way that the difference between two successive cost-to-go function and the utility function is minimum. In other words, to minimize

$$\sum_k [J(k) - \gamma J(k+1) - U(k)]^2.$$

Applying the gradient decent, the weights can be updated as follows

$$W_{critic}(k+1) = W_{critic}(k) + \Delta W_{critic}$$

$$\Delta W_{critic} = \alpha_c [J(k) - \gamma J(k+1) - U(k)] \frac{\partial J(k)}{\partial W_{critic}}$$

where $W_{critic}$ is the critic network weight and $\alpha_c$ is the learning rate.

B. *Action neural network*

The goal of action neural network is to generate the control signal to minimize the cost-to-go function for the immediate future. In other words, the objective is to minimize the sum of the utility function over the specific horizon, which is defined in this project as one second. ($1000\ ms$). The implemented action neural network is similar to that of critic network, a fully connected multi-layer feedforward neural network with two hidden layers, each with eight nodes, and one output. The input signal to this network is similar to the input of the critic neural network except it does not include the control signal. The action network input vector can be written as follows:

$$IN_{\{action\ network\}} = [P\ Q\ e_p\ e_q\ e_f\ \theta_i].$$

where $IN_{\{action\ network\}}$ is the input vector to the proposed action neural network. The output signal of the action network is the peak value of the output voltage of the inverter ($E(k)$). In order to update weights in the action network, the backpropagation algorithm is used. The objective is to minimize $J(k)$ as follows:

$$\zeta = \sum_k \frac{\partial J(k+1)}{\partial E(k)}$$

where $W_{action}$ is the action network weight and $\alpha_a$ is the learning rate. Consequently, the weights can be computed as follows:

$$\Delta W_{action} = -\alpha_a \zeta \frac{\partial \zeta}{\partial W_{action}}$$

$$W_{action}(k+1) = W_{action}(k) + \Delta W_{action}.$$

## IV. PERFORMANCE EVALUATION OF THE TRAINED HDP VIRTUAL INERTIA–BASED CONTROLLER

Figure. 6 illustrates the block diagram of the proposed HDP VSG controller. As shown, in the proposed controller the conventional PI controller is replaced with the HDP neural network controller. In favor of generating the proper input to the neural networks, a block named "network input generator" is implemented and by feeding the raw data to this block, it generates the compatible input vector.

In order to train the neural network the following steps should be followed.

1. A random vector for the initial states is generated
2. A random setting for the active and the reactive power is set.
3. The neural networks weights were initialized randomly.
4. Weights continued to be updated for the horizon of 1 sec.

This simulation analyzes the performance of the proposed controller in both inductive and resistive grids. The system parameters are listed in Table I.

### A. Inductive grid

Figure. 7 illustrates the comparison between the HDP controller and the conventional (PI) controller for a VSG connected to an inductive grid. As mentioned in Section II, the conventional PI performance in this operating point is acceptable because the delivered active power and reactive power are proportional to the inverter phase angle and the inverter voltage magnitude, respectively. However, with relatively similar active power overshoot for both HDP and PI controller, the HDP controller tends to reach steady state faster.

TABLE I
SYSTEM PARAMETERS USED IN SIMULATION

| Parameter | Value | Unit |
|---|---|---|
| DC voltage | 250 | V |
| AC line voltage | 110 | V |
| AC frequency | 60 | Hz |
| Moment of inertia | 0.1 | Kg.m2 |
| Frequency droop | %4 | -- |
| Inverter power rating | 5 | kW |
| *Inductive line* | | |
| Filter inductance | 1 | μH |
| Line inductance | 100 | μH |
| Line resistance | 10 | mΩ |
| *Resistive line* | | |
| Filter inductance | 1 | μH |
| Line inductance | 1 | μH |
| Line resistance | 500 | mΩ |
| *HDP parameters* | | |
| γ | 1 | |
| Sampling time | 1 | ms |
| [$\alpha_c\ \alpha_c$] | [1 1] | -- |
| [$k_P\ k_Q\ k_f$] | [1 1 0] | -- |

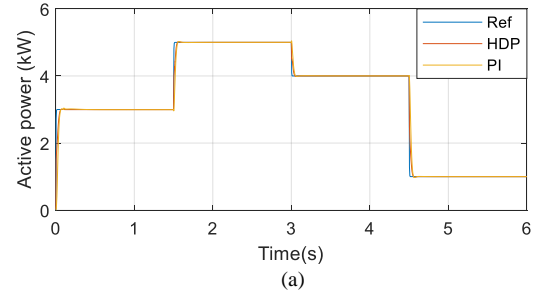
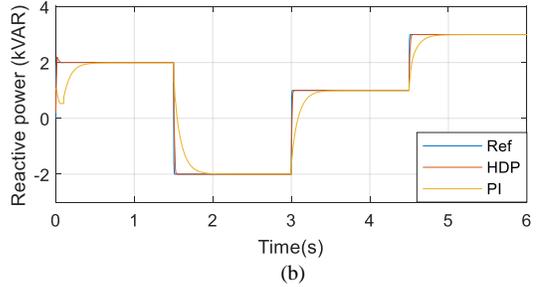
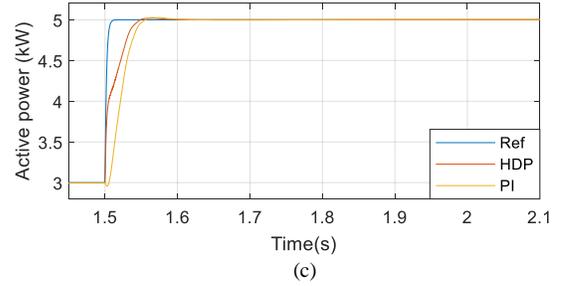
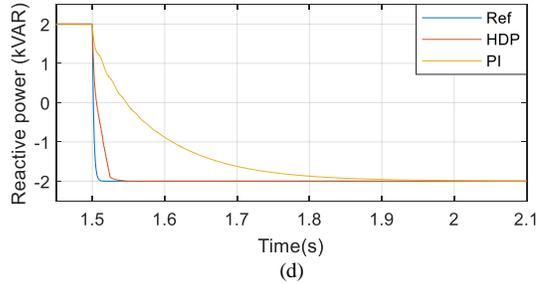

Figure. 7. HDP controller performance for VSG connected to the inductive grid (a) active power (b) active power (zoomed in) (c) reactive power (d) reactive power (zoomed in)

### A. Resistive grid

Figure. 8 shows the comparison between the HDP controller and the conventional PI controller in the resistive grid. It is discussed in Section II that the assumption that reactive and active power are proportional to the inverter voltage magnitude and phase angle, respectively, is no longer valid in resistive grids. Consequently, the conventional controller, which uses the reactive power error to regulate the inverter magnitude,

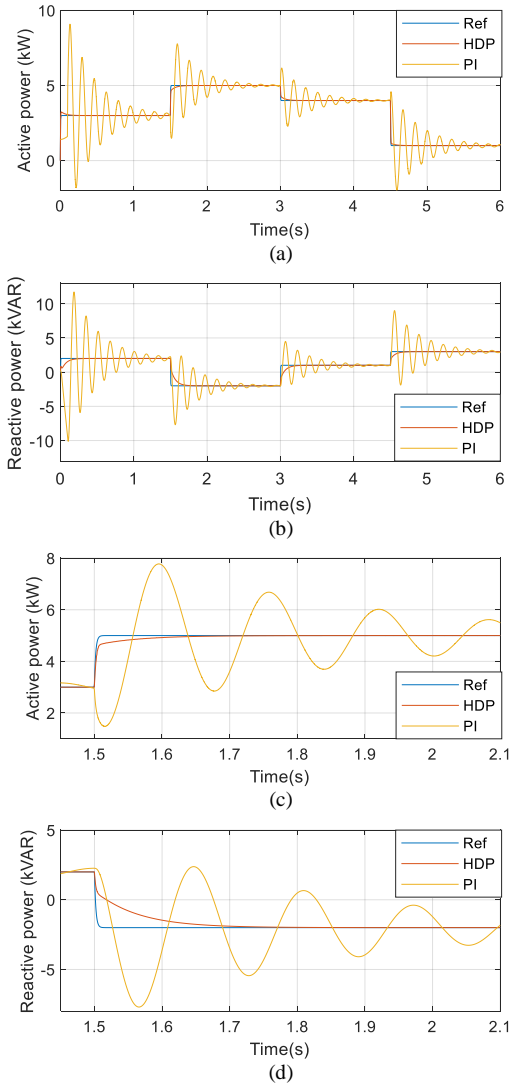

Figure. 8. HDP controller performance for VSG connected to the resistive grid (a) active power (b) active power (zoomed in) (c) reactive power (d) reactive power (zoomed in)

does not function properly. Nevertheless, the neural network–based nature of the HDP controller enables itself to adjust the networks' weights through online learning to guarantee the performance necessities.

## V. CONCLUSION

Three-phase inverters are widely implemented to connect RESs to the grid. Currently, these inverters are controlled with conventional PIs, which have several drawbacks such as lack of the ability to work in standalone mode and stability issues. To overcome these disadvantages, the VSG concept has emerged. The VSG control scheme is challenging, especially in non-inductive grids. This paper presents the heuristic dynamic programming controller to design a neural network–based adaptive controller for three-phase VSGs performing in inductive and resistive grids. It was shown by simulation that a HDP controller with well-trained action network and critic network, particularly in a resistive grid, performs better than the conventional PIs.

## VI. ACKNOWLEDGMENT

This material is based upon work supported by the U.S. Department of Energy, "Enabling Extreme Fast Charging with Energy Storage", DE-EE0008449.